\documentclass[10pt,twocolumn,letterpaper]{article}

\usepackage[pagenumbers]{cvpr} 

\newcommand{\modelname}{\textsc{OCO}}
\newcommand{\GRU}{\textsc{GRU}}

\newcommand{\SA}{\textsc{SA}}

\definecolor{first}{RGB}{200,0,0}   
\definecolor{second}{RGB}{0,153,0}  
\definecolor{third}{RGB}{0,0,200}  

\definecolor{header}{RGB}{60,60,60}  
\definecolor{lightgray}{RGB}{245,245,245}  
\definecolor{checkmark}{RGB}{0,150,0}  
\definecolor{xmark}{RGB}{180,0,0}  

\newcommand{\cmark}{\textcolor{checkmark}{\ding{51}}}
\newcommand{\xmark}{\textcolor{xmark}{\ding{55}}}

\definecolor{cvprblue}{rgb}{0.21,0.49,0.74}
\usepackage[pagebackref,breaklinks,colorlinks,allcolors=cvprblue]{hyperref}

\usepackage{bm}
\usepackage{colortbl} 
\usepackage{multirow}
\usepackage{float}
\usepackage[normalem]{ulem}
\usepackage[detect-all,separate-uncertainty = true]{siunitx}
\usepackage{colortbl} 
\usepackage{multirow}
\usepackage{booktabs}    
\usepackage{array}       
\usepackage{xcolor}
\usepackage{wasysym}
\usepackage{pifont} 
\usepackage[ruled,vlined]{algorithm2e}
\usepackage{graphicx}
\usepackage{amssymb}
\usepackage{float}
\usepackage{bbm}
\usepackage{amsthm}
\theoremstyle{definition}
\newtheorem{definition}{Definition}

\def\paperID{} 
\def\confName{CVPR}
\def\confYear{2026}

\AtBeginDocument{
  \crefformat{section}{\S#2#1#3}
}

\title{Mitigating Simplicity Bias in OOD Detection through Object \\ Co-occurrence Analysis}

\author{
Boyang Dai$^{1}$ \qquad
Chaoqi Chen$^{2}$ \qquad
Yizhou Yu$^{1,3}$\thanks{Corresponding author.}
\\
$^{1}$The University of Hong Kong 
$^{2}$Shenzhen University 
$^{3}$Shenzhen Loop Area Institute \\
{\tt\small boyangdai@connect.hku.hk, 
cqchen1994@gmail.com, 
yizhouy@acm.org}
}

\begin{document}
\maketitle
\begin{abstract}
Out-of-distribution (OOD) detection is crucial for ensuring the reliability of deep learning models.
Existing methods mostly focus on regular entangled representations to discriminate in-distribution (ID) and OOD data, neglecting the rich contextual information within images. 
This issue is particularly challenging for detecting near-OOD, as models with simplicity bias struggle to learn discriminative features in disentangled representations.  
The human visual system can use the co-occurrence of objects in the natural environment to facilitate scene understanding.
Inspired by this, we propose an Object-Centric OOD detection framework that learns to capture \textbf{O}bject \textbf{CO}-occurrence (OCO) patterns within images. 
The proposed method introduces a new OOD detection paradigm that understands object co-occurrence within an image by predicting disentangled representations for the test sample, then adaptively divides patterns into three scenarios based on object co-occurrence patterns observed in ID training data, and finally performs OOD detection in a divide-and-conquer manner.
By doing so, \modelname~can distinguish near-OOD by considering the semantic contextual relationships present in their images, avoiding the tendency to focus solely on simple, easily learnable regions.
We evaluate OCO through experiments across challenging and full-spectrum OOD settings, demonstrating competitive results and confirming its ability to address both semantic and covariate shifts. 
Code is released at \url{https://github.com/Michael-McQueen/OCO}.
\end{abstract}    
\vspace{-2mm}
\section{Introduction}
\label{sec:intro}
Deep learning has achieved remarkable success across various domains over the past decade~\cite{He_2016_CVPR, deng2019arcface}.  
Despite deep learning models exhibiting extraordinary performance on in-distribution (ID) data, they face significant challenges when distributional shifts occur~\cite{torralba2011unbiased, DBLP:conf/iclr/HendrycksG17}.  
To solve this issue, out-of-distribution (OOD) detection has been extensively investigated, aiming to make deep models more reliable when deployed in non-stationary test environments. 

Existing OOD detection methods predominantly rely on the distinctive characteristics between ID and OOD data in the latent feature space~\cite{lee2018simple, DBLP:conf/icml/SunM0L22KNN}, logit outputs~\cite{DBLP:conf/iclr/HendrycksG17, DBLP:conf/nips/LiuWOL20}, or a combination of both~\cite{wang2022vim, nnguide, she}. 
Although effective for far-OOD scenarios where distribution shifts are drastic, it often falters in near-OOD settings, where subtle semantic deviations make differentiation more challenging~\cite{DBLP:conf/nips/FortRL21, fang2022out, survey, ssb,yang2025oodd,chen2025dual,zhang2025adaptive,isaac2025fever,dong2025confound,dong2024adversarially}. 

\begin{figure}[!t]
    \centering
    \includegraphics[width=1.0\linewidth]{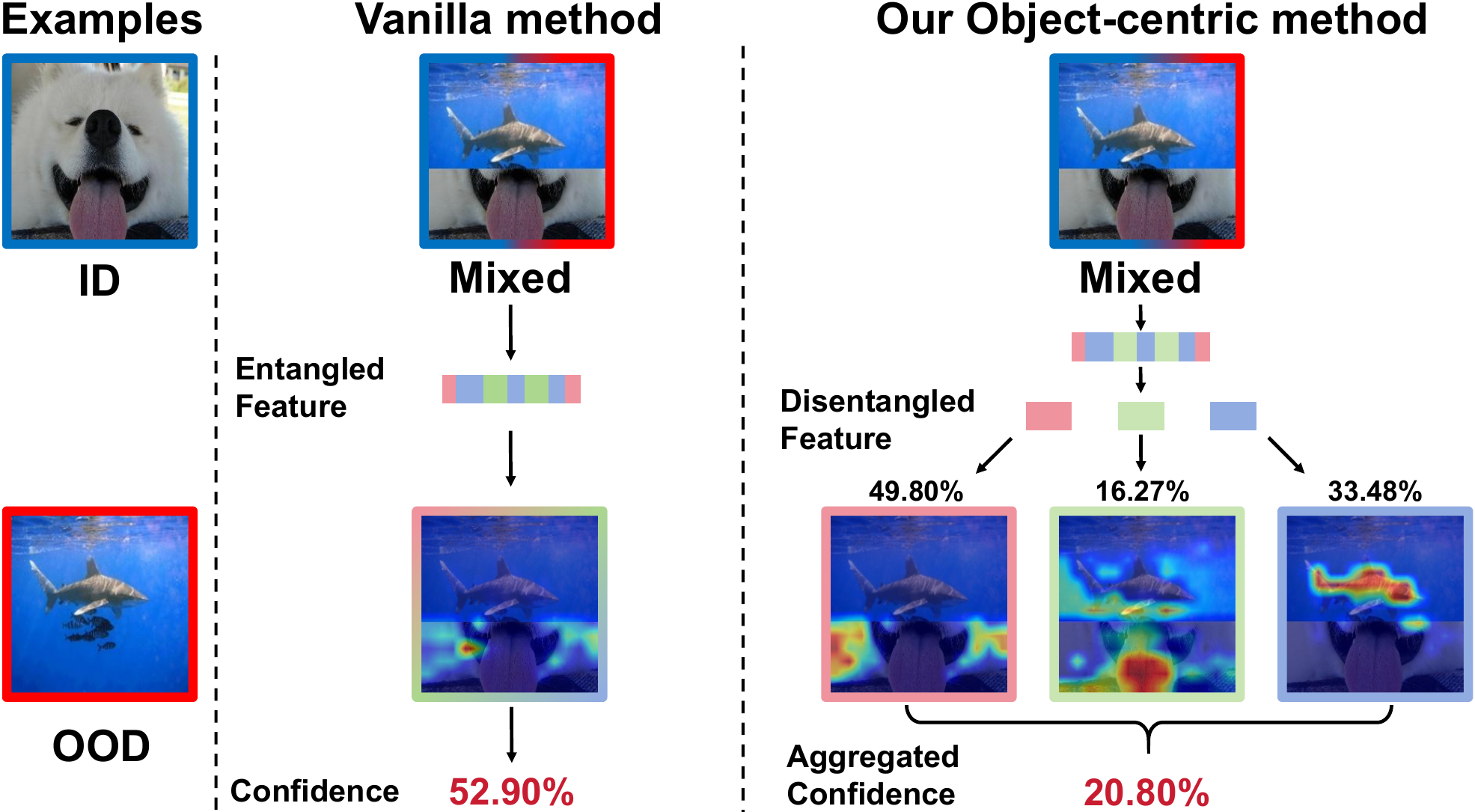}
    \caption{Attention visualization of vanilla method and object-centric prediction paradigms on mixed anomalous samples, \texttt{dog} is the ID (blue box), and \texttt{shark} is the OOD (red box).}
    \vspace{-3mm}
    \label{fig:teaser}
    \vspace{-1mm}
\end{figure}

This limitation stems from the models' simplicity bias -- when processing entangled representations, 
deep networks tend to prioritize easily learnable local cues while overlooking complex semantic relationships that are crucial for scene understanding~\cite{DBLP:conf/nips/ShahTR0N20, DBLP:conf/icml/TiwariS23, DBLP:conf/icml/ZhangZCJ024, chen2022compound, tian2019traffic, dong2026allies}. 
As shown in Fig.~\ref{fig:teaser}, 
for the anomalous image mixing ID and OOD samples, 
depicting an oceanic scene where a \texttt{dog} (ID) and a \texttt{shark} (OOD) co-occur. 
For the conventional approach, simplicity bias causes predictions based on the entangled feature\footnote{This refers to the predictive features from the model's final layer.} to focus solely on the dog's discriminative parts, 
neglecting the scene's contextual incongruity. 
This results in overconfident predictions, with a confidence of 52.90\%.

In contrast, humans can easily recognize the semantic incompatibility between a dog and a shark co-occurring in an oceanic setting.
Cognitive science research suggests that the human visual system leverages object relationships within scenes to make predictions about the world~\cite{s1, s2, s3}.
This gap between human and machine perception raises a critical question: \textit{How can object co-occurrence patterns be leveraged to mitigate simplicity bias in OOD detection?}

Recent studies in object-centric representation learning show that an image's entangled features can be decomposed into compositions of distinct objects~\cite{p1, p2, p3}.
Motivated by this insight, we argue that an effective OOD detection method should focus on the underlying compositions and patterns within an image, rather than treating it as a unified entity when distinguishing between ID and OOD data.

In response, we propose \modelname, a novel OOD detection framework that addresses simplicity bias through object co-occurrence (OCO) modeling. We utilize Slot Attention to abstract disentangled representations into slots that capture object co-occurrence. These slots are then aligned with semantics to form OCO patterns, which mirror human perceptual logic.
As shown in Fig.~\ref{fig:teaser} (right), each slot specializes in specific visual concepts—such as facial features, anatomical details, and contextual elements (e.g., ocean scene in our example)—while aggregated predictions help uncover semantic contextual incompatibility, reducing overconfidence from 52.90\% to 20.80\% in ambiguous cases.

For OOD detection, we propose more effective detection scores by incorporating object co-occurrence pattern information. Based on the nature of test samples' co-occurrence patterns, we categorize all possible patterns into three scenarios: (1) single patterns, (2) typical patterns, and (3) atypical patterns. Adopting this divide-and-conquer approach, rather than a one-size-fits-all paradigm, we devise tailored OOD detection scores for each scenario based on its distinct characteristics.

The main contributions are summarized as follows:



\begin{itemize}
    \item We propose OCO, a novel object-centric framework for OOD detection that leverages object co-occurrence patterns to learn complex scene-object semantic correlations, effectively mitigating the simplicity bias inherent in traditional models.

    \item Unlike methods that use a unified OOD score for all data, OCO adaptively categorizes test samples into different groups based on their co-occurrence patterns, assigning specific OOD scores to each group.  

    \item We demonstrate that OCO achieves competitive performance against state-of-the-art baselines on both challenging OOD and Full-spectrum OOD~\cite{DBLP:journals/ijcv/YangZL23} benchmarks, showing its ability to handle both semantic and covariate shifts effectively.

  
    
\end{itemize}

\begin{figure*}[!tbp]
    \centering
    \vspace{-2mm}
    \includegraphics[width=1\linewidth]{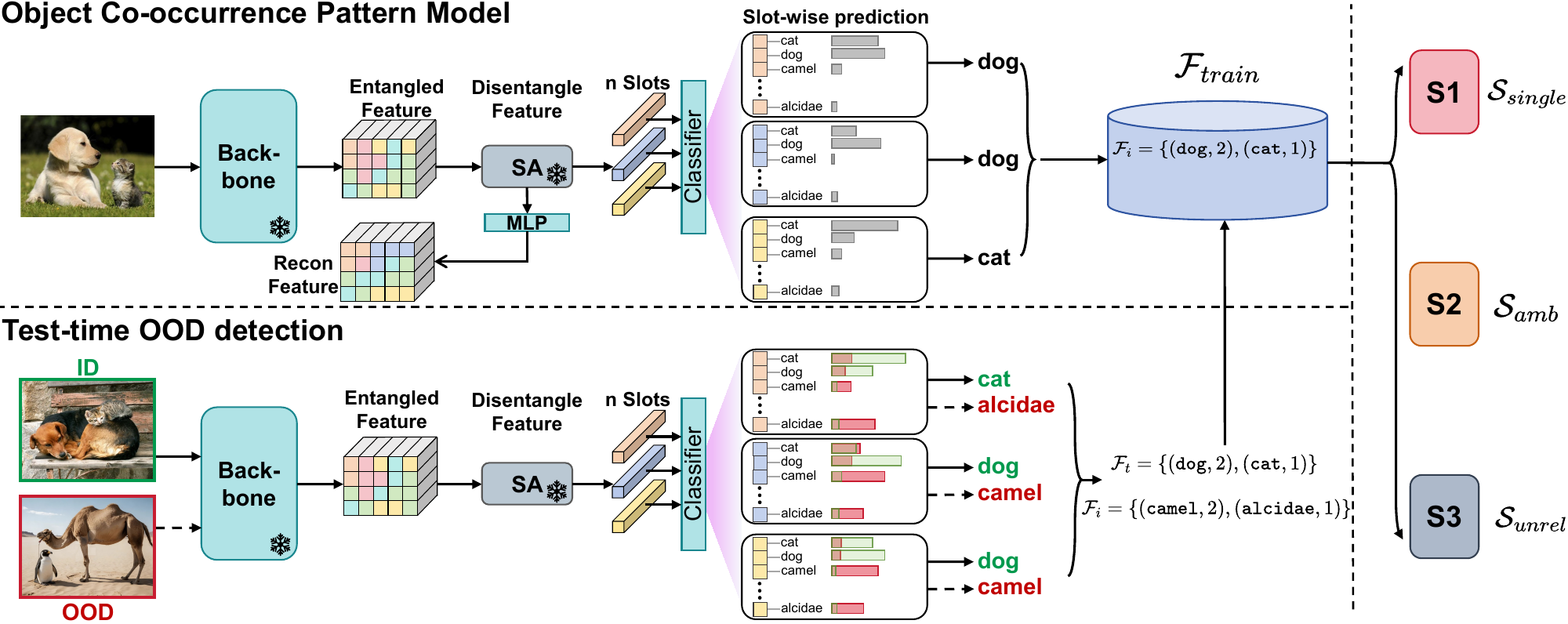}
    \caption{Overview of our \modelname. We first establish ID training data object co-occurrence pattern statistics ($\mathcal{F}_{train}$). During OOD detection, patterns are grouped according to $\mathcal{F}_t$ co-occurrences to compute OOD scores. The slot-wise prediction can learn object co-occurrence patterns. In the above example, \texttt{dog}, \texttt{cat}, \texttt{camel}, and \texttt{alcidae} are ID, while \texttt{penguin} is OOD, where \texttt{alcidae} and \texttt{penguins} have very similar semantic features.}
    \label{fig:pipeline}
    \vspace{-2mm}
\end{figure*}

\section{Preliminary}
\subsection{Preliminary: Slot Attention}
\label{m1}
Slot attention (SA)~\cite{slots} has emerged as a mainstream object-centric method that can iteratively extract object features into several slots in an unsupervised manner. Given an input feature map $\bm{x} \in \mathbb{R}^{H \times W \times C}$, where $H$ and $W$ are the size of the feature map, and $C$ is the feature map dimension. In \SA, the query $\bm{Q}$ is generated by sampling $K$ slots $\bm{S}\in \mathbb{R}^{K \times d}$ from a learnable Gaussian distribution, where $d$ is the dimension of each slot. The input feature map $\bm{x}$ is then transformed into the key $\bm{K}$ and value $\bm{V}$, which together with $\bm{Q}$ enable the cross-attention mechanism~\cite{vaswani2017attention} that competitively abstracts object features into slots:
\begin{equation*}
\bm{Q}^{(t)}=\bm{S}^{(t)}\bm{W}_Q, \qquad \bm{K}=\bm{xW}_K, \qquad \bm{V}=\bm{xW}_V,
\end{equation*}
$\bm{W}_Q$, $\bm{W}_K$, and $\bm{W}_V$ are learnable projection matrices.
Then, slot features are iteratively refined through a \GRU~\cite{DBLP:conf/ssst/ChoMBB14} that serves as a memory-like update mechanism:
\begin{equation*}
    \bm{A}^{(t)} = \textit{softmax}(\frac{\bm{Q}\bm{K}^T}{\sqrt{D}}), \quad \bm{S}^{(t+1)} = \text{GRU}(\bm{S}^{(t)}, \bm{A}^{(t)}\bm{V}).
\end{equation*}
\subsection{Problem Setup}
Let $\mathcal{D}_{\text{train}} = \{(\bm{X}_i, y_i)\}_{i=1}^n$ be a set of training instances sampled independently and identically from the source distribution $\mathcal{P}_{\text{in}}$, where $y_i \in \{1,\ldots,M\}$ denotes the corresponding class label. In the context of OOD detection, a model is first trained on $\mathcal{D}_{\text{train}}$, then deployed to identify samples drawn from an unknown target distribution $\mathcal{P}_{\text{out}}$. The detector $\mathcal{G}$ can be formally defined as:
\begin{equation}
\nonumber
    \mathcal{G}(\bm{X}, \tau) = \begin{cases}
        \text{inlier} & \text{if } \mathcal{S}(\bm{X}) \geq \tau \\
        \text{outlier} & \text{if } \mathcal{S}(\bm{X}) < \tau
    \end{cases}
\end{equation}
Here, $\mathcal{S}(\bm{X})$ denotes the detection function and $\tau$ represents a decision threshold calibrated to maintain a predetermined true positive rate (typically 95\%) on inlier samples.
\section{Method}
\label{method}
In this section, we propose \modelname, a novel object-centric OOD detection framework shown in Fig~\ref{fig:pipeline}. Subsequently, we first describe how to collect the object co-occurrence pattern in training data (\cref{m1}). Then, we analyze the object co-occurrence pattern between training data and test data (\cref{m2}). Finally, we propose specialized OOD detection scores based on the object co-occurrence patterns (\cref{m3}).

\subsection{Object Co-occurrence Pattern Modeling }
\label{m1}
\textbf{Disentangled Representation.} Given an entangled feature $\bm{x}_i\in\mathbb{R}^{H\times W\times D}$, we first leverage the off-the-shelf disentanglement representation method, specifically slot attention (SA)~\cite{slots}, to abstract the disentangled object feature into $K$ slots $\bm{S}_i=\{\bm{s}_i^{(1)}, \bm{s}_i^{(2)}, \dots, \bm{s}_i^{(K)}\} = SA(\bm{x}_i)\in\mathbb{R}^{K\times d}$, where each slot captures a distinct object part. 
To align slots with class-specific semantics, the classifier $h(\cdot; \theta)$ takes $K$ slots and returns corresponding $K$ outputs:   
\begin{equation}
\bm{l}_i^{(k)} = h(\bm{s}_i^{(k)}; \theta), \quad k = 1,2,\dots,K.
\label{eq: slots logits}
\end{equation}
where $\theta$ denotes the learnable parameters. Then these $K$ slot predictions are aggregated into a global prediction $\bm{l}_i = \sum_{k=1}^{K} \bm{l}_i^{(k)}$. 
The classification loss $
\mathcal{L}_{ce} = -\sum_{i=1}^{N} \log p_i(y_i)$ is computed using cross-entropy between the predicted class distribution $p_i$, derived from the softmax of $\bm{l}_i$, and the ground-truth label $y_i$.
To ensure the object-centric representation capability during semantic alignment, we introduce an auxiliary reconstruction loss $\mathcal{L}_{aux} = \|\bm{x}_i - \hat{\bm{x}_i}\|_2$ that minimizes the discrepancy between slot-reconstructed features $\hat{\bm{x}_i}=upsample(MLP(\bm{S}_i))$ and the entangled features $\bm{x_i}$.
The overall optimization objective combines the classification loss and the auxiliary reconstruction loss:
\begin{equation}
\mathcal{L} = \mathcal{L}_{ce} + \mathcal{L}_{aux}.
\label{loss}
\end{equation}
\textbf{Co-occurrence Discovery}. 
Through this object-centric model,  slots learn \textit{object co-occurrence patterns} within images, in which each slot independently computes logits that can act as a localized evidence provider and allow all slots' contributions to work together toward an accurate global prediction. 
\textit{e.g.} compatible combinations like dog and grass are reinforced by $\mathcal{L}_{ce}$, while anomalous pairings such as dog and ocean are suppressed.
This resembles an ensemble learning~\cite{DBLP:conf/iclr/HavasiJFLSLDT21}, where diverse weak slot predictors collaborate to form robust predictions.

To discover the overall object co-occurrence patterns, a systematic approach is employed to collect co-occurrence statistics from the ID training data.
Specifically, the process begins with an object-centric classifier that generates predictions for each training image $i$ across $K$ slots.
The slot-wise predictions $\mathcal{C}_i = \{c_i^{(k)}\}_{k=1}^K$ by applying argmax to slot-wise logits $c_i^{(k)} = \text{argmax}(\bm{l}_i^{(k)})$ in Eq.~\eqref{eq: slots logits}, where each $c_i^{(k)} \in \{1,2,...,M\}$ and $M$ denotes the total number of classes. 
To collect co-occurrence patterns, the first task is to identify the distinct object categories present in $\mathcal{C}_i$. This is done by constructing the unique categories set $\mathcal{U}_i$:
\begin{equation}
\mathcal{U}_i = \{c \mid \exists k \in \{1,\dots,K\}, c_i^{(k)} = c\}.
\end{equation}
This step eliminates duplicates and focuses on which distinct objects are detected in the image.
We note that slot-based approaches may suffer from over-segmentation, where a single larger object might be represented across multiple slots. To obtain more robust patterns, we calculate the frequency of each unique category in $\mathcal{C}_i$. This is represented in the frequency set $\mathcal{F}_i$:
\begin{equation}
   \mathcal{F}_i = \left\{\left(c, \sum_{k=1}^K \mathbb{I}(c_i^{(k)} = c)\right) \,\bigg|\, c \in \mathcal{U}_i \right\},
\end{equation}
where $\mathbb{I}(c_i^{(k)} = c)$ is the indicator function that equals $1$ when $c_i^{(k)} = c$ and $0$ otherwise.

To formally identify when multiple object categories appear together in an image, we define a co-occurrence configuration as follows:
\begin{definition}[\textbf{Co-occurrence Configuration}]
\label{def:cooccur}
\textit{For image $\bm{x}_i$, with slot predictions unique categories set $\mathcal{U}_i$, let $\mathcal{F}_i$ be the frequency set, we define its co-occurrence configuration: 
\begin{itemize}  
    \item \textbf{Single-category Configuration}: When $|\mathcal{F}_i| = 1$, where all slots converge to one category  
    \item \textbf{Multi-category Configuration}: When $|\mathcal{F}_i| \geq 2$, where presence of multiple distinct categories.
\end{itemize}}
\end{definition}
Finally, to understand co-occurrence patterns across the training dataset, only the frequency sets $\mathcal{F}_i$ from training instances where a co-occurrence event occurs (\emph{i.e.,} $|\mathcal{F}_i| \ge 2$) are aggregated into a comprehensive collection $\mathcal{F}_{train}$:
\begin{equation}
\mathcal{F}_{train} = \bigcup_{i \in D_{train}} \mathcal\{{F}_i\},
\end{equation}
where $D_{train}$ denotes the set of all training instances.
Each element in $\mathcal{F}_{train}$ is a distinct frequency set representing a specific combination of object categories and their frequencies, as seen in at least one training image. This collection reflects the scene context present in the training data.

By incorporating object co-occurrence frequencies $\mathcal{F}_i$, \modelname~ achieves three benefits: (1) enhanced robustness against spurious predictions while capturing complex semantic relationships, (2) avoiding oversimplified feature representations, 
and (3) preventing coincidental associations when collecting ID co-occurrence pattern statistics.
\subsection{Test-time Co-occurrence Pattern Division}
\label{m2}
During the testing phase, suppose we have a test sample with an object co-occurrence pattern frequency set denoted as $\mathcal{F}_t$. When facing OOD samples, since the model is only trained on ID data, it tends to predict the closest ID category it knows. This can create unusual category combinations unlike patterns seen during training. According to definition \ref{def:cooccur}, we categorize this sample into the following scenario:
\begin{itemize}
    \item \textbf{Scenario 1 (S1)} (\textit{Single Patterns}): The frequency set $\mathcal{F}_t$ contains only one element, indicating that all slot categories $\{c_t^{(k)}\}_{k=1}^K$ are the same class, resulting in cardinality $|\mathcal{F}_t| = 1$. For example, $\mathcal{F}_t = \{(\texttt{cat}, 3)\}$ shows that all three slots consistently predict \texttt{cat}.

    \item \textbf{Scenario 2 (S2)} (\textit{Typical Patterns}): Multi-object pattern matches training observations. Formally, $|\mathcal{F}_t| \ge 2 \wedge \mathcal{F}_t \in \mathcal{F}_{train}$. For example, $\mathcal{F}_t = \{(\texttt{dog}, 2), (\texttt{cat}, 1)\} \in \mathcal{F}_{train}$ represents a common scene with \texttt{dog} and \texttt{cat}. 

    \item \textbf{Scenario 3 (S3)} (\textit{Atypical Patterns}): Multi-object pattern mismatches training observations. Formally,  $|\mathcal{F}_t| \ge 2  \wedge \mathcal{F}_t \notin \mathcal{F}_{train}$. For example, $\mathcal{F}_t = \{(\texttt{penguin}, 2), (\texttt{camel}, 1)\} \notin \mathcal{F}_{train}$ represents an unlikely combination of objects that rarely co-exist in natural scenarios.
\end{itemize}
This semantic relationship is consistent with human perception.
This way of categorizing provides insights into whether a test sample's co-occurrence patterns align with the training distribution or deviate in a manner indicative of OOD characteristics.

\begin{figure}[t]
    \centering
    \includegraphics[width=0.85\linewidth]{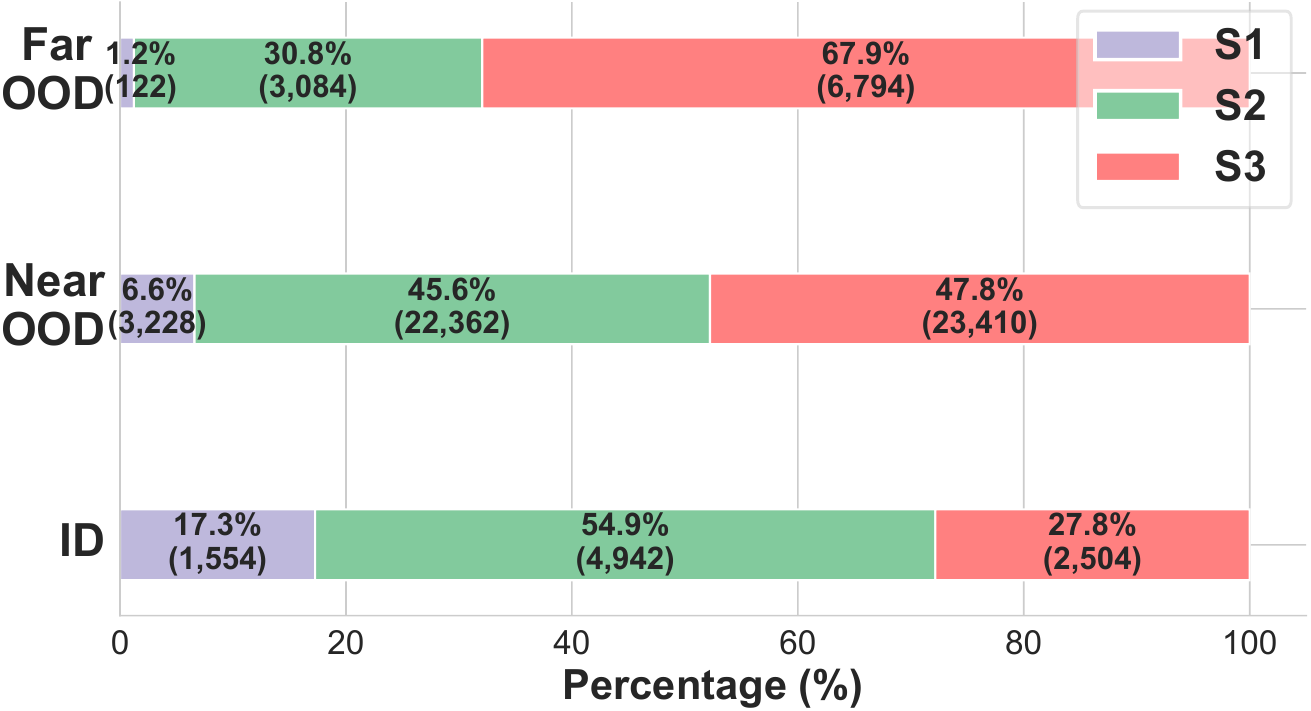}
    \caption{Number of samples in each group for ID (ImageNet-200), Near-OOD (SSB-hard), and Far-OOD (iNaturalist) datasets.}
    \vspace{-4mm}
    \label{fig:group_num}
    \vspace{-2mm}
\end{figure}

To offer a deeper understanding and rationale for partitioning test samples via object co-occurrence patterns, we analyze the sample distribution across partitions. As shown in Fig.~\ref{fig:group_num}, we observe that ID test data predominantly reside in S2 (54.9\%), confirming our method's fidelity to normal compositions. The higher proportion of near-OOD in S2 compared to far-OOD stems from their part-level feature similarity to ID data, whereas far-OOD exhibits a dominant S3 presence (67.9\%) due to fundamentally implausible combinations. The clear difference demonstrates that object co-occurrence patterns reliably differentiate between subtle near-OOD variations and far-OOD anomalies.
\subsection{OOD Detection with OCO}
\label{m3}      

As mentioned before, there exist three scenarios. First, we analyze the distinctive characteristics between ID and OOD under each scenario and design a corresponding score to detect OOD.

For \textbf{S1}, when all object slots consistently predict the same category, the main challenge in OOD detection lies in the \textit{overestimation problem}~\cite{DBLP:conf/icml/GuoPSW17,DBLP:conf/icml/WeiXCF0L22}. We mitigate this through dual-confidence calibration. We use scene-level confidence, which is the aggregated prediction certainty $P_t = \max_c(\text{softmax}(\bm{l}_t))$, and object-level confidence, which measures the highest slot agreement $p_{t}^{\text{max}} = \max_{k}( \max_{c}(\text{softmax}(\bm{l}_t^{(k)})))$.
The calibrated score combines both evidences:  
\begin{equation}
    \mathcal{S}_{\text{single}} = P_t \cdot p_{t}^{\text{max}}.
    \label{g1}
\end{equation}

\textbf{S2} exhibits multiple predicted categories with co-occurrence patterns known from $\mathcal{F}_{train}$. Such test samples are highly \textit{ambiguous} between ID and OOD, especially in near-OOD cases~\cite{DBLP:conf/iccv/FanLLW023}. To solve this issue, we propose an uncertainty-aware scoring mechanism based on Dempster–Shafer theory (DST)~\cite{dempster, shafer}. Unlike traditional probability methods, DST enables more flexible uncertainty modeling, especially with conflicting evidence (for details see Supp. Section 7). In our score, maximum probability $p^{\text{max}}$ represents the highest confidence in a category's existence, capturing the most significant evidence while calculating pairwise belief contradictions between the dominant category $c'$ and others $c \in \mathcal{F}_t$.
Belief Combination:  
\begin{equation}
    \text{Bel}(c', c) =\underbrace{p_{c'}^{\text{max}}p_{c}^{\text{max}}}_{\text{Pattern Likelihood}} + \underbrace{p_{c'}^{\text{max}}(1-p_{c}^{\text{max}}) + (1-p_{c'}^{\text{max}})p_{c}^{\text{max}}}_{\text{Ambiguous Evidence}}.
\end{equation}
The ``Pattern Likelihood'' represents expected co-occurrence probability, and ``Ambiguous Evidence'' indicates a sample that must belong to one of the ID classes, though it's uncertain which specific class, and definitely not OOD. 
The aggregated uncertainty score across all slot predictions:  
\begin{equation}
    \mathcal{S}_{\text{amb}} = \frac{1}{|\mathcal{F}_t|-1} \sum_{c \neq c'} \text{Bel}(c', c).
    \label{g2}
\end{equation}

\textbf{S3} comprises samples with previously unseen co-occurrence patterns. Due to these atypical patterns, most slot-wise predictions $\bm{l}^{(k)}_t$ become \textit{unreliable}. We therefore directly quantify anomaly by an object-level confidence score, defined as the highest slot prediction:  
\begin{equation}
\mathcal{S}_{\text{unrel}} = p_{t}^{\text{max}} = \max_{k,c} (\text{softmax}(\bm{l}_t^{(k)})).
\label{g3}
\end{equation}

\section{Experiments}
\label{sec:exp}
\begin{table*}[!htbp]
    \centering
    \renewcommand{\arraystretch}{1.3} 
    \setlength{\tabcolsep}{1pt}
    \resizebox{0.9\textwidth}{!}{
\begin{tabular}{clcccccc}
\toprule
 & \multicolumn{7}{c}{\textbf{OOD Datasets}}\\
       \cmidrule(lr){3-8} 
 \multirow{1}{*}{\textbf{Model}} & \multirow{1}{*}{\textbf{Methods}}&\textbf{SSB-hard} & \textbf{NINCO} & \textbf{iNaturalist} & \textbf{Texture} & \textbf{OpenImage-O} & \textbf{Mean} \\
 & & FPR95$\bm{/}$AUROC & FPR95$\bm{/}$AUROC & FPR95$\bm{/}$AUROC & FPR95$\bm{/}$AUROC & FPR95$\bm{/}$AUROC & FPR95$\bm{/}$AUROC\\
\midrule
 \multirow{9}{*}{ViT} & Energy ~\cite{DBLP:conf/nips/LiuWOL20}& 84.54 / 67.29 & 
67.39 / 79.17 & 
27.52 / 92.61 & 
60.53 / 83.92 & 
47.49 / 86.89 & 57.49 / 81.98\\
  & MaxLogit ~\cite{maxlogit} &
84.13 / 67.78 & 
66.19 / 79.92 & 
26.73 / 93.06 & 
60.26 / 84.13 & 
46.92 / 87.42 & 56.85 / 82.46\\

  & SHE ~\cite{she} &  
89.97 / 59.99 & 
82.23 / 71.16 & 
58.92 / 84.17 & 
84.83 / 74.21 & 
67.34 / 81.42 & 76.66 / 74.19\\
 &NNguide~\cite{nnguide} & 
85.54 / 62.02 & 
75.05 / 70.65 & 
42.98 / 87.00 & 
69.21 / 79.80 & 
56.43 / 82.04 & 65.84 / 76.30\\
 &  SCALE ~\cite{Scale} &  
84.42 / 67.79 & 
66.54 / 79.75 & 
23.92 / 93.53 & 
55.50 / 84.85 & 
43.81 / 87.90 & 54.84 / 82.76\\
&NECO ~\cite{neco} & 
86.53 / 70.27 & 
83.04 / 79.28 & 
33.23 / 92.66 & 
59.43 / 87.37 & 
67.24 / 86.44 & 65.89 / 83.20\\
& FDBD~\cite{fdbd} & 
88.50 / 64.31 & 
68.90 / 79.15 & 
18.22 / 95.33 & 
36.53 / 90.19 & 
36.17 / 90.73 & 49.66 / \underline{83.94} \\
& CoRP ~\cite{corp}& 
88.73 / 60.07 & 
70.42 / 72.63 & 
28.26 / 93.52 & 
11.28 / 97.12 & 
44.55 / 86.71 & 
\underline{48.65} / 82.01\\
& OODD~\cite{yang2025oodd} & 
84.34 / 72.05 & 
67.52 / 77.12 & 
30.41 / 91.51 & 
38.63 / 88.31 & 
52.35 / 90.54 & 
54.65 / 83.91 \\
\rowcolor{gray!30}
 & OCO (Ours) &  
76.82 / 73.21 & 
60.77 / 81.84 & 
30.31 / 92.41 & 
35.66 / 91.54 & 
32.76 / 91.22 & 
\textbf{47.26} / \textbf{86.04} \\
\midrule
\midrule 
\multirow{9}{*}{DINOv2} &  Energy ~\cite{DBLP:conf/nips/LiuWOL20} &  
83.58 / 63.62 & 
64.65 / 75.78 & 
14.24 / 95.46 & 
71.71 / 79.61 & 
39.53 / 88.39 & 54.74 / 80.57\\
 & MaxLogit ~\cite{maxlogit} & 
83.16 / 64.53 & 
63.75 / 77.08 & 
13.79 / 95.79 & 
71.99 / 80.09 & 
38.74 / 88.96 & 54.29 / 81.29\\
 & SHE ~\cite{she} & 
79.28 / 68.90 & 
62.23 / 79.76 & 
31.10 / 92.75 & 
74.42 / 77.85 & 
49.84 / 86.01 & 59.37 / 81.05\\
 & NNguide ~\cite{nnguide} & 
87.67 / 57.15 & 
75.43 / 67.77 & 
24.83 / 89.96 & 
67.27 / 78.44 & 
51.48 / 83.46 & 61.34 / 75.36\\
 & SCALE ~\cite{Scale} &  
83.50 / 63.42 & 
63.98 / 75.83 & 
13.40 / 95.73 & 
69.23 / 79.88 & 
37.87 / 88.86 & 53.60 / 80.74\\

&  NECO ~\cite{neco} &  
83.63 / 64.17 & 
63.64 / 77.19 & 
12.67 / 96.12 & 
67.45 / 81.65 & 
37.06 / 89.66 & 52.89 / 81.76\\

 & FDBD~\cite{fdbd} & 
84.21 / 65.45 & 
54.41 / 81.52 & 
11.99 / 96.43 & 
37.25 / 90.61 & 
30.30 / 91.89 & 43.63 / 85.18\\
& CoRP ~\cite{corp}& 
84.99 / 62.44 & 
59.36 / 80.31 & 
16.81 / 95.81 & 
12.64 / 96.54 & 
28.86 / 93.26 & 
\underline{40.53} / 85.67\\
& OODD~\cite{yang2025oodd} & 
80.12 / 76.22 & 
58.68 / 81.30 & 
15.49 / 93.78 & 
29.10 / 92.36 & 
31.25 / 91.54 & 
42.93 / \underline{87.04} \\
\rowcolor{gray!30}
 & OCO (Ours) & 
78.43 / 72.13 & 
57.74 / 81.27 & 
10.38 / 97.45 & 
18.36 / 94.36 & 
28.55 / 93.53 & 
\textbf{38.70} / \textbf{87.75} \\   
\bottomrule
\end{tabular}
 }
\caption{OOD detection results trained on ImageNet-1k and tested on 5 OOD datasets. The utilized metrics include FPR95 (\(\downarrow\)), aiming for lower values to indicate better performance; AUROC (\(\uparrow\)), where higher values denote superior discriminative ability. The top-performing methods are marked with \textbf{bold} for the best and \underline{underline} for the second best.}
    \label{ood}
\vspace{-3mm}
\end{table*}

\begin{table*}[!htbp]
    \centering
\renewcommand{\arraystretch}{1.3} 
    \setlength{\tabcolsep}{1pt}
    \resizebox{0.9\textwidth}{!}{
\begin{tabular}{clcccccc}
    \toprule
    & \multicolumn{7}{c}{OOD Datasets}\\
       \cmidrule(lr){3-8} 
     \multirow{1}{*}{\textbf{Model}} & \multirow{1}{*}{\textbf{Methods}} & \textbf{SSB-hard} & \textbf{NINCO} & \textbf{iNaturalist} & \textbf{Texture} & \textbf{OpenImage-O} & \textbf{Mean} \\
    & & FPR95/AUROC & FPR95/AUROC & FPR95/AUROC & FPR95/AUROC & FPR95/AUROC & FPR95/AUROC \\
    \midrule
    \multirow{9}{*}{ViT} &Energy ~\cite{DBLP:conf/nips/LiuWOL20} & 
90.28 / 51.83 & 
    78.94 / 63.50 & 
    48.61 / 80.43 & 
    74.23 / 69.31 & 
    64.89 / 72.42 & 71.39 / 67.50
    \\
    & MaxLogit ~\cite{maxlogit} &  
90.03 / 52.30 & 
78.15 / 64.50 & 
    47.75 / 81.34 & 
    74.07 / 69.63 & 
    64.50 / 73.30 & 70.90 / 68.21
    \\
   &SHE ~\cite{she} &
93.57 / 47.63 & 
    88.36 / 58.78 & 
    71.73 / 73.09 & 
    90.14 / 62.96 & 
    77.84 / 69.96 & 84.33 / 62.48
    \\
    &NNguide ~\cite{nnguide} &
    90.93 / 49.57 & 
    83.90 / 58.25 & 
    59.54 / 76.90 & 
    79.89 / 68.92 & 
    70.46 / 70.95 & 
    76.94 / 64.92 \\
    &SCALE ~\cite{Scale} &
90.28 / 51.94 & 
    78.49 / 63.58 & 
    45.76 / 81.22 & 
    70.88 / 69.45 & 
    62.41 / 72.93 & 69.56 / 67.82
    \\
    & NECO ~\cite{neco} &
90.95 / 51.41 & 
    78.74 / 64.07 & 
    47.47 / 81.85 & 
    69.57 / 72.08 & 
    62.91 / 74.14 & 69.93 / 68.71
   \\
   & FDBD~\cite{fdbd} & 
92.73 / 49.80 & 
    79.58 / 64.18 & 
    39.01 / 85.37 & 
    55.53 / 77.10 & 
    55.21 / 77.91 & 64.41 / 70.87
    \\
    & CoRP ~\cite{corp}& 
    92.73 / 48.71 & 
    80.04 / 61.02 & 
    40.10 / 84.34 & 
    46.28 / 84.64 & 
    60.30 / 78.18 & 
    \underline{63.89} / \underline{71.38} \\
& OODD ~\cite{yang2025oodd} &  
84.56 / 73.02 & 
70.35 / 59.13 & 
54.45 / 75.22 & 
58.21 / 69.15 & 
53.12 / 71.98 & 
64.14 / 69.70 \\ 
    \rowcolor{gray!30}
    &\modelname~(Ours) &
84.93 / 57.97 & 
74.19 / 66.99 & 
49.69 / 81.34 & 
54.82 / 80.67 & 
52.37 / 79.09 & 
\textbf{63.20} / \textbf{73.21}
    \\
    \midrule
    \midrule 
   \multirow{9}{*}{DINOv2} & Energy ~\cite{DBLP:conf/nips/LiuWOL20} &
89.60 / 48.04 & 
    77.00 / 59.53 & 
    33.84 / 85.60 & 
    81.83 / 65.03 & 
    58.25 / 74.51 & 68.10 / 66.54
    \\
    & MaxLogit ~\cite{maxlogit} &
89.35 / 49.05 & 
    76.42 / 61.16 & 
    32.99 / 86.53 & 
    82.01 / 65.91 & 
    57.49 / 75.64 & 67.65 / 67.66
    \\
    &SHE ~\cite{she} &
94.98 / 40.65 & 
    92.94 / 48.23 & 
    74.72 / 64.77 & 
    91.10 / 56.71 & 
    82.59 / 61.58 & 
    87.27 / 54.39
   \\
    &NNguide ~\cite{nnguide} &
92.18 / 43.33 & 
    84.14 / 52.53 & 
    44.57 / 77.38 & 
    78.60 / 64.79 & 
    67.31 / 69.88 & 73.36 / 61.58
    \\
   &SCALE ~\cite{Scale} &
89.70 / 47.54 & 
    76.75 / 59.17 & 
    33.23 / 85.78 & 
    80.32 / 64.76 & 
    57.15 / 74.66 & 67.43 / 66.38
   \\
    &NECO ~\cite{neco} &
89.74 / 48.54 & 
    76.48 / 60.99 & 
    31.96 / 86.90 & 
    79.05 / 67.41 & 
    56.37 / 76.43 & 66.72 / 68.05
   \\
    &FDBD~\cite{fdbd} & 
90.34 / 49.46 & 
    70.15 / 65.26 & 
    31.54 / 87.59 & 
    57.11 / 79.00 & 
    51.21 / 79.56 & 60.07 / 72.17
    \\
    & CoRP ~\cite{corp}& 
    91.05 / 46.22 & 
    74.37 / 64.23 & 
    29.70 / 88.34 & 
    43.83 / 85.45 & 
    49.56 / 82.65 & 57.70 / 73.38\\
& OODD ~\cite{yang2025oodd} &  
80.50 / 54.88 & 
78.45 / 61.60 & 
41.53 / 79.00 & 
44.10 / 88.32 & 
42.92 / 86.20 & 
\underline{57.50} / \underline{74.00} \\
    \rowcolor{gray!30}
    &\modelname~(Ours) & 
85.07 / 58.25 & 
71.41 / 67.48 & 
28.03 / 90.75 & 
47.44 / 86.58 & 
46.29 / 83.92 & 
\textbf{55.65} / \textbf{77.40}
    \\   
    \bottomrule
    \end{tabular}
 }
    \caption{
    FS-OOD detection results trained on ImageNet-1k, tested on 5 OOD datasets and 3 covariate shift ID datasets (ImageNet-v2/C/R).
    }
    \label{fs-ood}
    \vspace{-3mm}
\end{table*}

\subsection{Experimental Setup}
\textbf{Datasets.} 
For comprehensive experiments, we adopt the OpenOOD benchmark~\cite{yang2022openood, zhang2023openood}, which provides an accurate, and standardized evaluation framework. We use \textit{ImageNet-1k}~\cite{ImageNet} as our ID dataset. For OOD evaluation, we introduce both near-OOD datasets (\underline{SSB-hard}~\cite{ssb} and \underline{NINCO}~\cite{ninco}) and far-OOD datasets (\underline{iNaturalist}~\cite{van2018inaturalist}, \underline{Textures}~\cite{textures}, and \underline{OpenImage-O}~\cite{wang2022vim}). For thorough comparisons, we further extend our \modelname~to a challenging full-spectrum OOD (FS-OOD) evaluation~\cite{DBLP:journals/ijcv/YangZL23} that includes both covariate-shifted ID datasets (\textit{ImageNet-v2/C/R})~\cite{imagenetv2, imagenetc, imagenetr} and semantic-shifted OOD datasets mentioned above. Note that, we conduct all ablation studies and analyses on ImageNet-200, which consists of the first 200 classes of ImageNet-1K. See Supp. Section 8.1 for detailed descriptions.\\
\textbf{Baselines.}
We compare \modelname~with 9 OOD detection baselines: \textbf{Energy}\cite{DBLP:conf/nips/LiuWOL20}, \textbf{MaxLogit}\cite{maxlogit}, \textbf{SHE}\cite{she}, \textbf{NNguide}\cite{nnguide}, \textbf{SCALE}\cite{Scale}, \textbf{NECO}\cite{neco}, \textbf{FDBD}~\cite{fdbd},  \textbf{CoRP}~\cite{corp}, and
\textbf{OODD}~\cite{yang2025oodd}. 
\\
\textbf{Evaluation metrics.}
We evaluate our method using (1) the false positive rate (FPR95) at the threshold where the true positive rate for ID samples is 95\% and (2) the area under the receiver operating characteristic curve (AUROC). 
Both metrics are reported as percentages. 
In the ablation study, FPR95 and AUROC are averaged across all datasets.
\\
\textbf{Training details.}
We fine-tuned a single-layer linear classification head for 20 epochs on both pre-trained ImageNet-1k ViT-B/16~\cite{vit} and DINOv2 ViT-B/14~\cite{dino} models with AdamW~\cite{DBLP:conf/iclr/LoshchilovH19} optimizer using a learning rate of $0.0004$ and a cosine learning rate decay schedule that gradually reduces the learning rate from the initial value to $0.00005$. For object-centric slot attention, we adopted the pre-trained DINOSAUR~\cite{dinosaur} architecture. For a fair comparison, all baselines were evaluated on fine-tuned models. The complete training details can be found in Supp. Section 8.3.
\subsection{Main Results}
\textbf{OOD Detection.} The comparison results between \modelname~and other competitive baselines are presented in \cref{ood}. \modelname~consistently achieves competitive results across OOD datasets using both supervised ViT and self-supervised DINOv2 backbones. 
In particular, the proposed OCO outperforms OODD~\cite{yang2025oodd} by 2.13\% and 0.71\% in terms of AUROC on ViT and DINOv2 backbones, respectively.
Notably, we observe the following phenomena: (1) \modelname~achieves better overall performance with DINOv2 backbone compared to ViT, especially on the iNaturalist dataset. This is attributed to the self-supervised DINOv2 backbone's stronger feature disentanglement capability~\cite{dino, DBLP:conf/eccv/WangZS24}, leading to more reliable co-occurrence patterns in slot predictions. (2) \modelname~exhibits exceptional performance on SSB-hard dataset, a challenging near-OOD benchmark that is difficult to detect due to fine-grained category differences between ID and OOD. This demonstrates the effectiveness of leveraging co-occurrence patterns in \modelname.
\\
\textbf{FS-OOD Detection.}
\cref{fs-ood} compares the performance of \modelname~with current leading baselines on FS-OOD. Our method outperforms OODD~\cite{yang2025oodd}, achieving improvements of 3.51\% and 3.40\% in AUROC for the ViT and DINOv2 backbones, respectively. Particularly, utilizing the superior DINOv2 backbone, \modelname~excels on the most challenging SSB-hard dataset, enhancing AUROC by 3.37\% compared to the runner-up method. While most alternative approaches struggle on the SSB-hard dataset when using DINOv2 backbone, our success highlights how the object co-occurrence-based divide and conquer score effectively handles both semantic and covariate shifts, demonstrating robust generalization capabilities.

\subsection{Ablation Study}
\begin{table}[!tbp]
\centering
\belowrulesep=0pt\aboverulesep=0pt
\renewcommand{\arraystretch}{1.3}
\resizebox{0.85\linewidth}{!}{
\begin{tabular}{c|>{\centering\arraybackslash}p{1.2cm}>{\centering\arraybackslash}p{1.2cm}|c|c}
\toprule
\multirow{2}{*}{\textbf{\#ID}} & \multirow{2}{*}{$\mathcal{L}_{aux}$} & \multirow{2}{*}{\textbf{OCO}}  & \textbf{OOD} & \textbf{FS-OOD}\\
& & & {FPR95}{/AUROC} & {FPR95} {/AUROC}\\
\midrule
1 & \xmark & \xmark  & 35.99 / 90.76 & 72.50 / 64.70\\ 
2 & \xmark & \cmark  & 41.77 / 89.43 & 74.70 / 63.61\\
3 & \cmark & \xmark & \underline{30.70} / \underline{92.77} & \underline{67.58} / \underline{67.74} \\
4 & \cmark & \cmark  & \textbf{28.23} / \textbf{93.69} & \textbf{65.72} / \textbf{69.56} \\
\bottomrule
\end{tabular}
}
\vspace{-1mm}
\caption[]{Ablation of the proposed method.}
\vspace{-4mm}
\label{ablation}
\end{table}
In \cref{ablation}, we conducted an ablation study on the reconstruction constraint $\mathcal{L}_{aux}$ in Eq.~\eqref{loss} and the \modelname~score. We observed the following phenomena: (1) Comparing the results of cases \#1 and \#2, we found that the \modelname~was ineffective when $\mathcal{L}_{aux}$ was absent. This is because, without $\mathcal{L}_{aux}$, the slots were unconstrained and lost the ability to extract object features, leading to the failure of utilizing the co-occurrence patterns of slots. As illustrated in Fig.~\ref{vis}, the absence of $\mathcal{L}_{aux}$ significantly reduces the object localization capability of SA, leading to under-segmentation issues.  Thereby affecting the prediction of object co-occurrence patterns and leading to the failure of the method.
(2) In cases \#3 and \#4, after ensuring the slots' ability to abstract object features by incorporating $\mathcal{L}_{aux}$. The effect of the \modelname~was significantly improved, outperforming the other cases.
\begin{figure}[!tbp]
	\centering
	\small
	\setlength\tabcolsep{4mm}
	\renewcommand\arraystretch{0.6}
	\begin{tabular}{ccc}
		\includegraphics[width=0.087\textwidth]{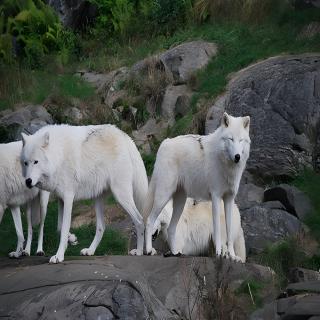} &
		\includegraphics[width=0.087\textwidth]{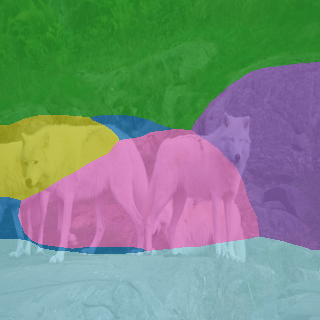} &
		\includegraphics[width=0.087\textwidth]{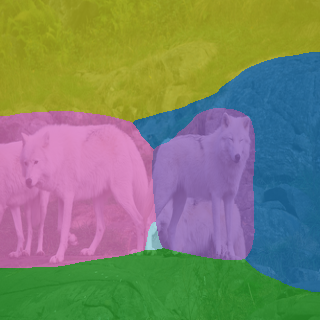} 
        \\
        \includegraphics[width=0.087\textwidth]{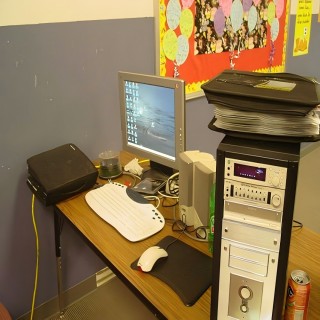} &
		\includegraphics[width=0.087\textwidth]{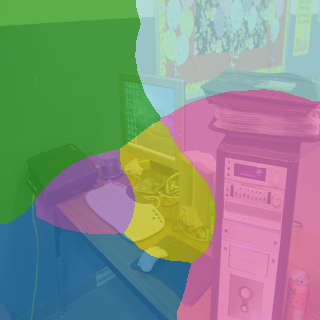} &
		\includegraphics[width=0.087\textwidth]{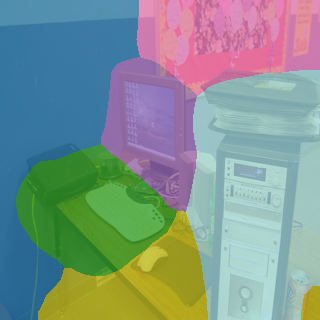} 
        \\
		(a) Raw image & (b) w/o $\mathcal{L}_{aux}$ & (c) w/ $\mathcal{L}_{aux}$ \\
	\end{tabular}
    \caption{Illustration of the raw image and segmentation masks with and without $\mathcal{L}_{aux}$ under 6 different slot numbers.}
    \label{vis}
    \vspace{-5mm}
\end{figure}

\begin{figure}
    \centering
    \includegraphics[width=0.91\linewidth]{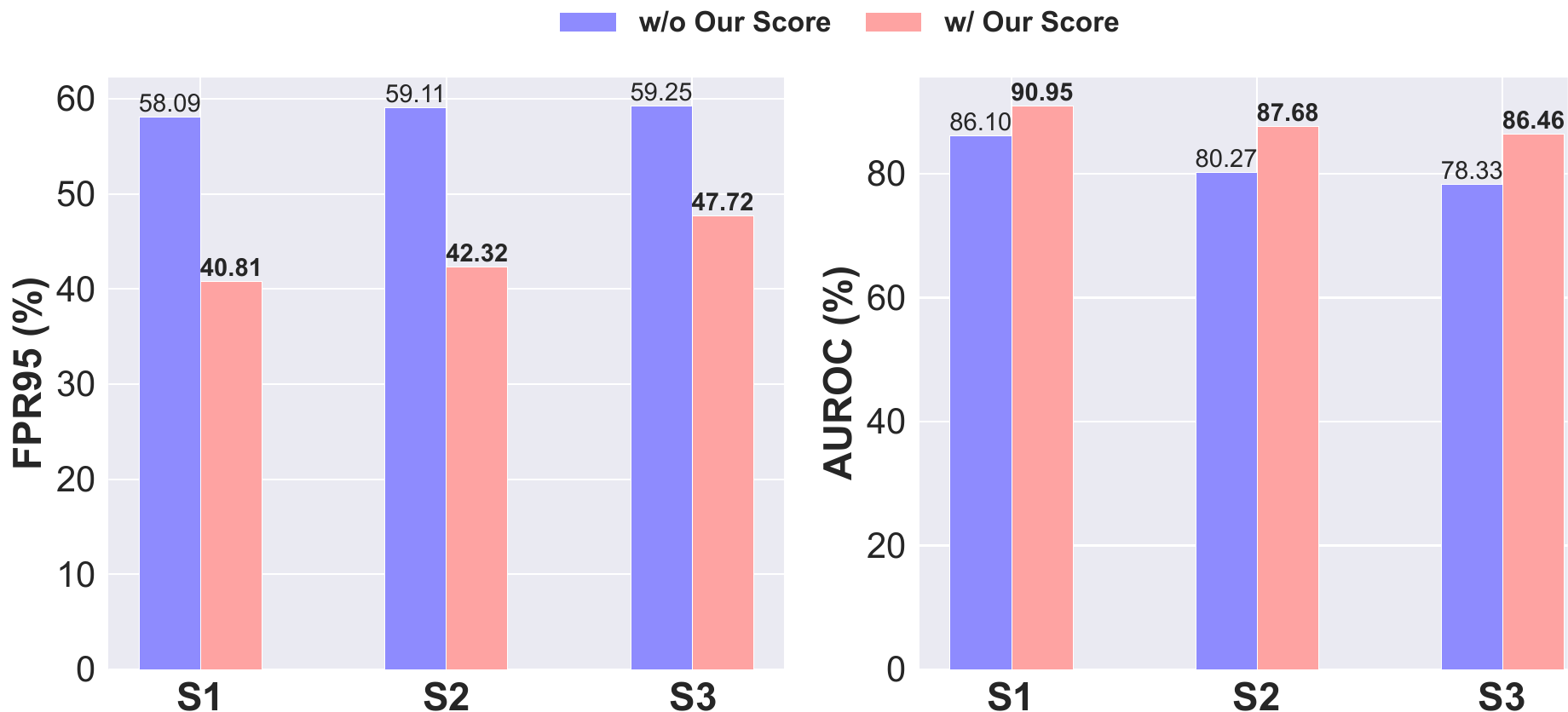}
    \caption{OOD detection results under different scenarios.}
    \label{group_ood}
    \vspace{-5mm}
\end{figure}
We further investigated the OOD detection performance under three different scenario conditions, with results shown in Fig.~\ref{group_ood}.
In \textbf{S1}, FPR95 decreased by 17.28\%, demonstrating the importance of mitigating overconfidence in our OOD detection score $\mathcal{S}_{single}$ (Eq.~\eqref{g1}). \textbf{S2} and \textbf{S3} exhibited significant improvements in both FPR95 and AUROC. FPR95 decreased by 16.79\% and 11.53\%, while AUROC increased by 7.41\% and 8.13\%, respectively. These results validate the effectiveness of the scores $\mathcal{S}_{amb}$ and $\mathcal{S}_{unrel}$ based on co-occurrence patterns (Eqs.~\eqref{g2} and~\eqref{g3}).
\subsection{Effect of Slot Count}
\begin{figure}[!tbp]
    \centering
    \includegraphics[width=0.75\linewidth]{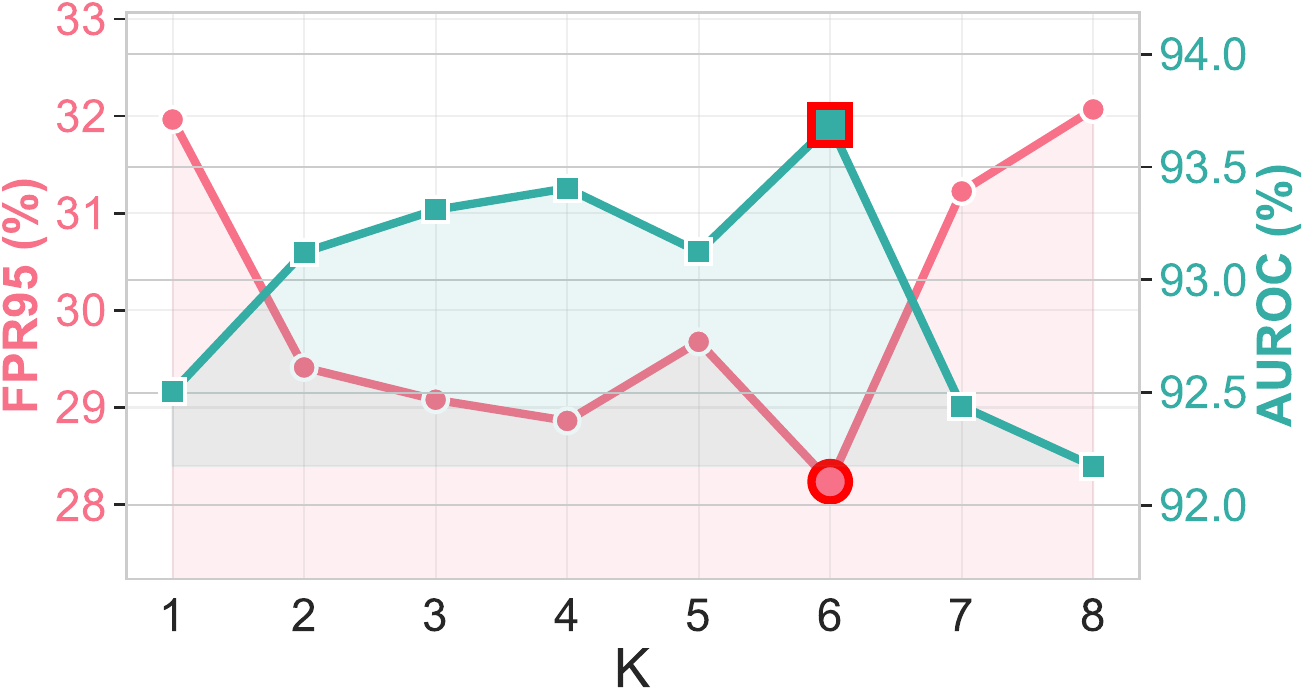}
    \vspace{-3mm}
    \caption{OOD detection results on different slot numbers.}
    \label{slot_num}
    \vspace{-5mm}
\end{figure}

Since this parameter $K$ affects $|\mathcal{C}_i|$ and consequently influences the co-occurrence patterns, on which our method heavily relies, the accuracy of these patterns is crucial. Fig.~\ref{slot_num} presents our results on the impact of slot numbers. Our experiments reveal that performance is suboptimal with fewer slots but gradually improves as the slot count increases. This improvement can be attributed to higher fault tolerance in contribution patterns within $\mathcal{C}_i$ with more slots. However, after reaching peak performance at $K=6$, further increasing the slot count leads to a sharp performance drop. This degradation occurs because excessive slots may over-segment object features, rendering the contribution patterns ineffective.

\subsection{Further Studying of Generalization}
To further investigate the role of co-occurrence patterns in improving generalization, we conducted experiments focusing on covariate shift classification. Differing from standard classification settings, we considered an object-centric setting, \textit{i.e.,} a sample to be correctly classified if the ground truth label appeared in $\mathcal{C}_i$, thereby examining the trustworthiness of co-occurrence patterns. As shown in \cref{acc}, on the standard ImageNet-200 dataset, the accuracy improved by 2.47\% compared to regular classification. Notably, on the challenging covariate shift datasets (ImageNet-v2/c/r), we observed significant accuracy improvements of 4.85\%, 6.10\%, and 7.86\%, respectively. These enhancements can be attributed to the strong generalization ability of co-occurrence patterns, leading to our superior performance in FS-OOD detection compared to other baselines.

\subsection{Quantitative Results}
\renewcommand{\arraystretch}{1.3}
\belowrulesep=0pt\aboverulesep=0pt

\begin{figure}[t]
	\centering
	\small
	\setlength\tabcolsep{4mm}
	\renewcommand\arraystretch{0.3}
	\begin{tabular}{cccc}
		\includegraphics[width=0.2\textwidth]{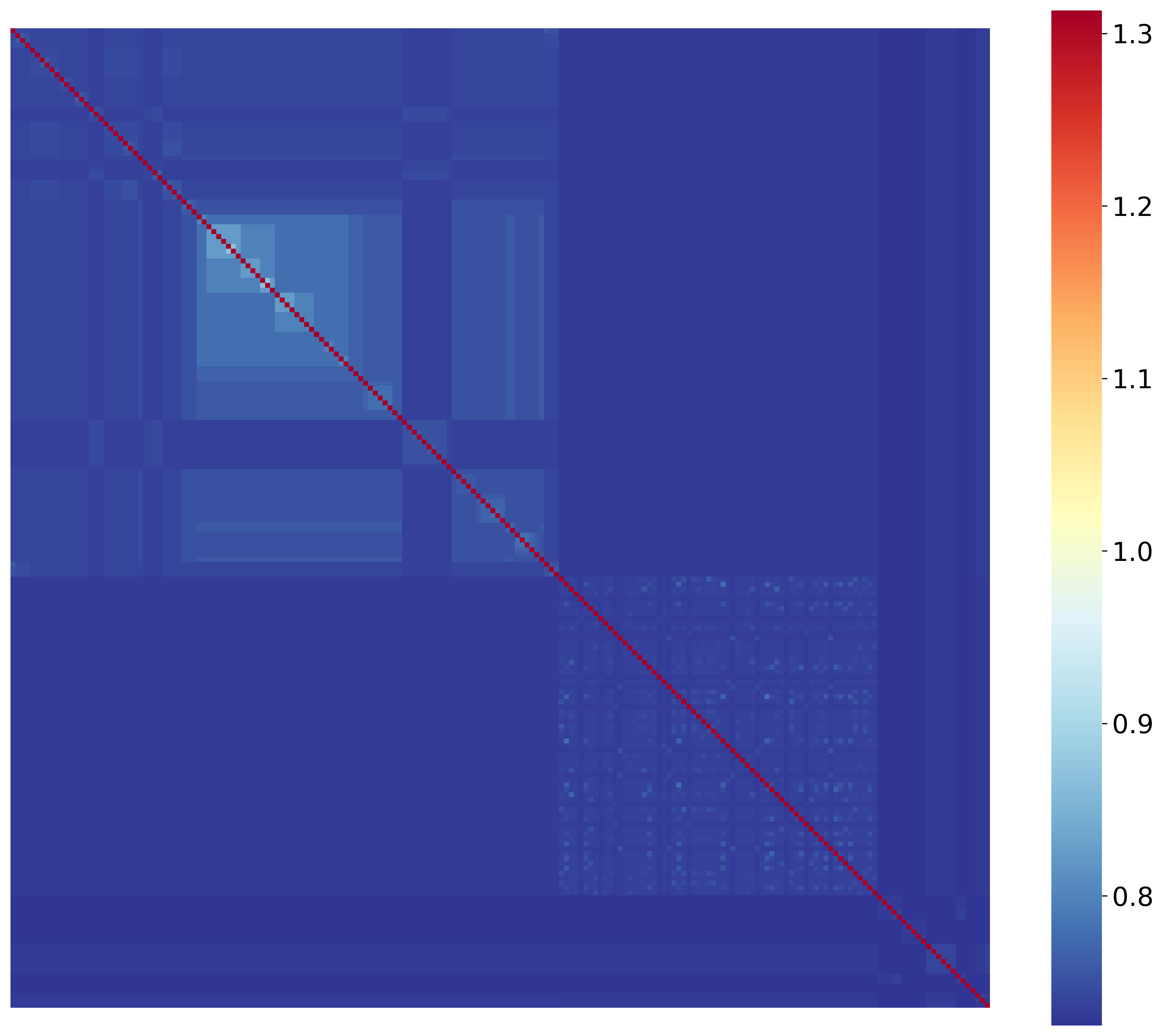} &
		\includegraphics[width=0.2\textwidth]{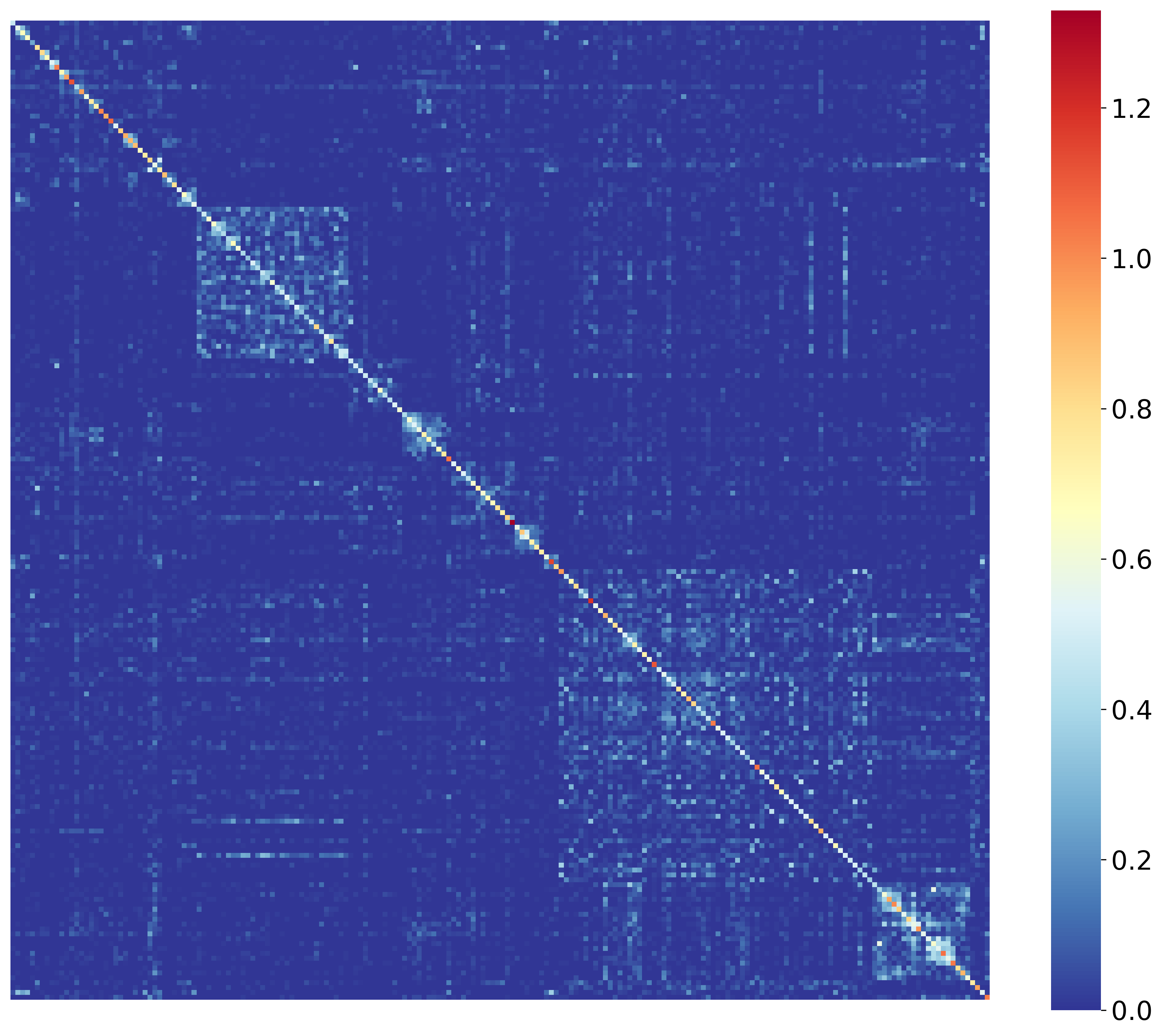} \\
		(a) WordNet & (b) ImageNet-200 \\
	\end{tabular}
    \caption{Visualization of object co-occurrence probabilities versus WordNet-based semantic similarities.}
    \label{vis_matrix}
\end{figure}

\begin{table}[!tbp]
    \centering
    \resizebox{0.95\linewidth}{!}{
    \begin{tabular}{l|>{\centering\arraybackslash}p{1.5cm}>{\centering\arraybackslash}p{1.5cm}>{\centering\arraybackslash}p{1.5cm}>
    {\centering\arraybackslash}p{1.5cm}}
        \toprule
        \multirow{2}{*}{\textbf{Method}}& \textbf{ImageNet-v2} & \textbf{ImageNet-c} & \textbf{ImageNet-r} & \textbf{ImageNet-200} \\
        \midrule
        Standard & 86.45 & 76.20 & 44.84& 94.04\\
        Object-centric & \textbf{91.30} & \textbf{82.30} & \textbf{52.70}& \textbf{96.51}\\ 
        \bottomrule
    \end{tabular}
    }
    \caption{The classification results of regular (ImageNet-200) and covariate shift dataset (ImageNet v2/c/r).}
    \label{acc}
    \vspace{-2mm}
\end{table}
The visualization of the object co-occurrence patterns matrix is shown in Fig.~\ref{vis_matrix}. We constructed a similarity matrix for each label using WordNet \cite{fellbaum1998wordnet}. Note that the WordNet similarity matrix serves as a sanity check for our co-occurrence patterns' semantic structure, not their accuracy. By comparing the object co-occurrence matrix with the similarity matrix, it is evident that the predicted object co-occurrence patterns matrix correctly expresses the accurate similarity relationships in most regions. This indicates that employing the object co-occurrence pattern for OOD detection effectively comprehends the semantic relationships within the image.
Our statistical analysis (Tab.~\ref{tab}) shows moderate alignment between predicted patterns and WordNet hierarchy, validating the model's ability to learn semantic relationships.

\section{Related Work}
\label{sec:rw}
\noindent
\textbf{OOD Detection} can be categorized into two main approaches: re-training and post-hoc. For re-training methods, OE~\cite{DBLP:conf/iclr/HendrycksMD19} introduces real outliers during training. VOS~\cite{DBLP:conf/iclr/DuWCL22VOS} and NPOS~\cite{DBLP:conf/iclr/TaoDZ023} synthesize virtual outliers
using multivariate Gaussian distributions.
Post-hoc methods include maximum softmax prediction (MSP)~\cite{DBLP:conf/iclr/HendrycksG17} and logit scores~\cite{maxlogit}. Energy-based transformations~\cite{DBLP:conf/nips/LiuWOL20} have been introduced to address the overconfidence issue. ReAct~\cite{react}, DICE~\cite{dice}, ASH~\cite{ash}, SHE~\cite{she} and NECO~\cite{neco} are inference-time enhancements, with NECO specifically leveraging neural collapse geometry and principal component analysis. CoRP~\cite{corp} employs non-linear mappings with cosine-Gaussian kernels to enhance out-of-distribution detection performance. However, these methods often struggle with near-OOD detection due to their inherent simplicity bias. In contrast to existing approaches, our method focuses on leveraging contextual co-occurrence patterns, utilizing the internal semantic structure for more robust OOD detection.
\\
\textbf{Object-Centric Learning} has gained increasing attention in computer vision. Early works like MONet~\cite{monet} and IODINE~\cite{iodine} propose to decompose scenes into object-centric representations through iterative inference. Recent advances like Slot Attention~\cite{slots} reformulate this procedure into a single encoding phase powered by iterative attention. 
Fan et al.~\cite{fan1} combine slot attention for object localization with unsupervised CLIP-based semantic extraction to achieve open-vocabulary object localization in videos.
EoRaS~\cite{fan2} leverages object-centric supervision and multi-view features to mutually enhance mask prediction in real-world scenarios. 
MESH~\cite{mesh} combines the computational efficiency of regularized optimal transport with the accuracy of unregularized transport to achieve better attention allocation.
OpenSlot~\cite{yin2024openslot} addresses mixed open-set recognition through object-centric learning, using slot features to represent diverse class semantics. 
In particular, DINOSAUR~\cite{dinosaur} introduces a transformer-based architecture that can effectively learn object-level features from real-world image datasets without object-level supervision. 
In comparison, our method primarily focuses on exploiting slot-wise predictions of object co-occurrence patterns to address the challenging task of OOD detection.

\begin{table}[]
  \centering
  \resizebox{0.6\linewidth}{!}{
  \begin{tabular}{lc}
  \toprule
  \textbf{Metric} & \textbf{Value} \\
  \midrule
  Spearman & 0.429 \\
  p-value & $1.549 \times 10^{-148}$ \\
  Cosine Similarity & 0.769 \\
  \bottomrule
  \end{tabular}}
  \caption{Statistical analysis between OCO and WordNet.}
  \label{tab}
  \vspace{-4mm}
\end{table}

\section{Conclusion}

In this work, we propose \modelname, an object-centric pipeline capable of capturing contextual information within the image through the learning of object co-occurrence patterns, and leverage this information to design a divide-and-conquer OOD scoring strategy for different object co-occurrence patterns. 
\modelname~achieves excellent performance on both semantic and covariate shift benchmarks, demonstrating its ability to address challenges from different types of test shifts. Our in-depth analysis provides insights into the effectiveness and generalization capabilities of object co-occurrence patterns.
Moreover, our approach is simple, extensible, and holds promise for real-world applications.

\clearpage

\clearpage
\setcounter{page}{1}
\maketitlesupplementary
\begin{figure*}[!htbp]
    \centering
    \includegraphics[width=0.8\linewidth]{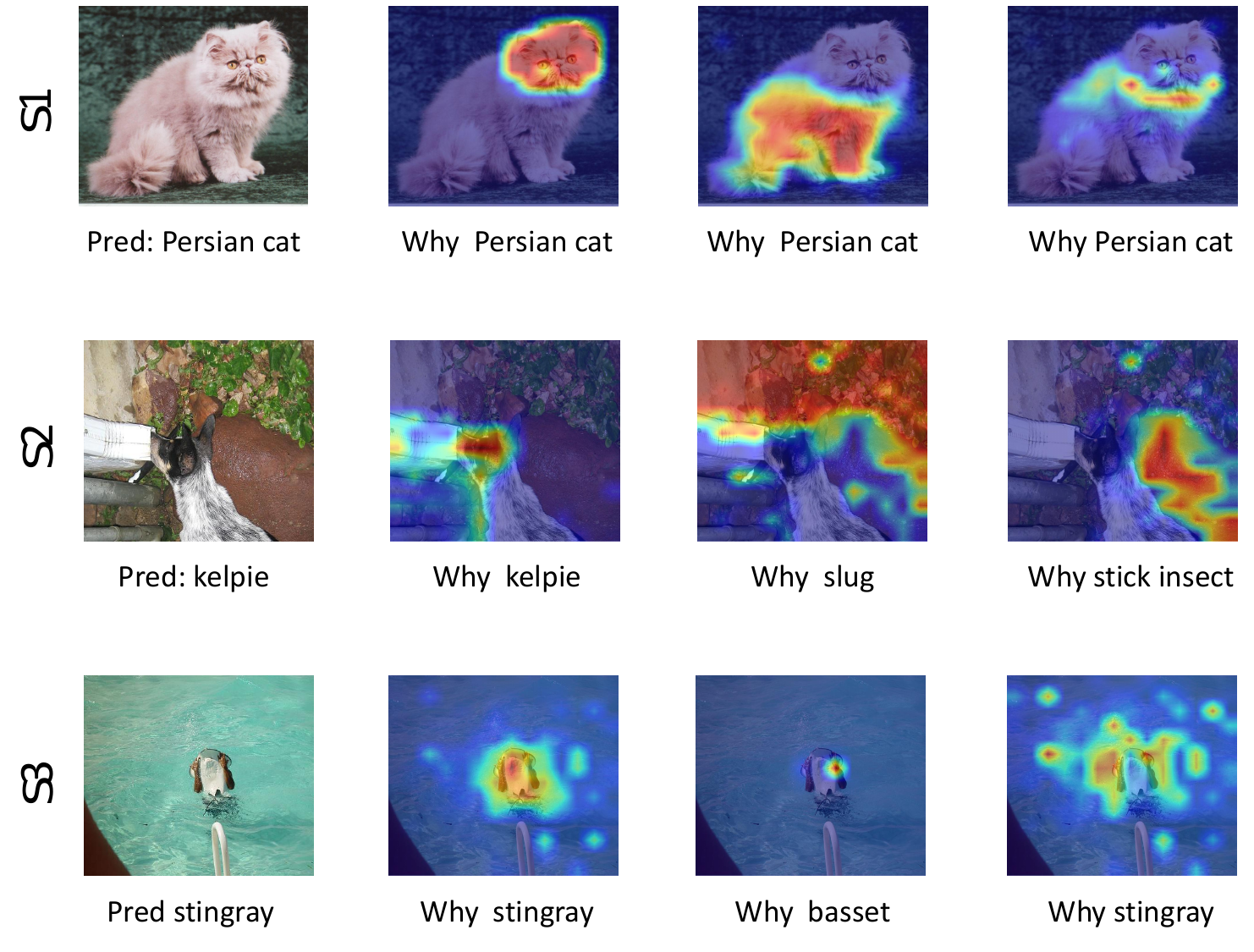}
    \caption{Attention visualization of ID.}
    \label{fig:id}
\end{figure*}
\section{Dempster-Shafer Theory}
Let $\Omega$ be a finite set called the frame of discernment, which represents all possible states or hypotheses in a given context. The power set of $\Omega$, denoted as $2^\Omega$, contains all possible subsets of $\Omega$ including the empty set $\emptyset$. A basic probability assignment (BPA) or mass function $m$ is a mapping from $2^\Omega$ to $[0,1]$ that satisfies:
\begin{equation}
    m(\emptyset) = 0 \text{ and } \sum_{\mathcal{A} \subseteq \Omega} m(\mathcal{A}) = 1
\end{equation}

For any subset $\mathcal{A} \subseteq  \Omega$, $m(\mathcal{A})$ represents the degree of evidence supporting exactly $\mathcal{A}$, not including any of its proper subsets. Based on the mass function, the belief function $Bel$ and plausibility function $Pl$ are defined as:
\begin{equation}
Bel(\mathcal{A}) = \sum_{\mathcal{B} \subseteq \mathcal{A}} m(\mathcal{B})
\end{equation}
\begin{equation}
Pl(\mathcal{A}) = \sum_{\mathcal{B} \cap \mathcal{A} \neq \emptyset} m(\mathcal{B})
\end{equation}
where $Bel(\mathcal{A})$ represents the total belief committed to $\mathcal{A}$ and all its subsets, while $Pl(\mathcal{A})$ measures the total belief that does not contradict $\mathcal{A}$. The interval $[Bel(\mathcal{A}), Pl(\mathcal{A})]$ can be interpreted as the lower and upper bounds of the probability of $\mathcal{A}$.
Given two pieces of evidence represented by mass functions $m_1$ and $m_2$ from independent sources, Dempster's rule of combination $\oplus$ defines their fusion as:
\begin{equation}
(m_1 \oplus m_2)(\mathcal{A}) = K^{-1} \sum_{\mathcal{B} \cap C = \mathcal{A}} m_1(\mathcal{B})m_2(\mathcal{C})
\end{equation}
where $K = 1 - \sum_{\mathcal{B}\cap \mathcal{C}=\emptyset} m_1(\mathcal{B})m_2(\mathcal{C})$ is the normalization factor, and $K \neq 0$. This combination rule provides a formal mechanism for evidence fusion and belief updating.

In \modelname, to reduce complexity, we only consider the categories predicted from aggregated features and slot predictions, rather than examining all $2^K$ possible combinations.
\section{Experiment}
\subsection{Datasets}
\textbf{SSB-hard} consists of 49,000 images covering 980 categories selected from ImageNet-21K. Classes outside ImageNet-1K but still within the ImageNet, making it semantically close and thus near OOD. \\
\textbf{NINCO} is a noise-free dataset of 5,879 images manually curated and verified by humans to ensure complete freedom from noise and contamination from ImageNet-1K classes. \\
\textbf{iNaturalist} offers natural world species images that differ significantly from ImageNet-1K's object categories.\\
\textbf{Textures} presents a fundamentally different challenge by focusing on textural patterns rather than object recognition.\\
\textbf{OpenImage-O} is carefully curated from the Open Images dataset, and provides diverse image content with 1,763 images specifically designated for validation.\\
\textbf{ImageNet-c}: it contains 15 corruption types (e.g., noise, blur, weather effects) each with 5 severity levels. The benchmark randomly samples 10K images across these 75 corruption-severity combinations for evaluation. \\
\textbf{ImageNet-r}: it tests generalization to artistic renditions, presenting ImageNet objects in various styles including sketches, paintings, and cartoons. \\
\textbf{ImageNet-v2}: it tests generalization under data collection bias. This dataset helps evaluate whether models are truly learning robust features or are overfitting to specific characteristics of the original ImageNet distribution. \\

\subsection{Baselines}
\textbf{Energy} is a post-processing method based on energy functions, which transforms the model's logits into energy scores to distinguish between ID and OOD samples. \\
\textbf{MaxLogit} utilizes the maximum logit value as the detection score, avoiding the smoothing effect introduced by the softmax function to achieve more discriminative OOD detection.\\
\textbf{SHE} integrates energy functions with feature prototypes, utilizing a hybrid approach that combines both energy-based scoring and prototype-based feature comparison. \\
\textbf{NNguide} presents a hybrid approach that combines energy-based scoring with feature-space nearest neighbor distance, leveraging both energy functions and feature similarities for detection. \\
\textbf{SCALE} implements a thresholded energy function approach that requires no access to training data, making it particularly practical for deployment scenarios. \\
\textbf{NECO} employs PCA reconstruction distance as its core mechanism for OOD detection, measuring the dissimilarity between original and reconstructed features. \\
\textbf{FDBD} identifies OOD samples by locating and leveraging the decision boundaries in the ID feature space, focusing on the geometric properties of the learned representations. \\
\textbf{CoRP} applies non-linear kernel methods to detect out-of-distribution data, using explicit feature mappings with cosine and cosine-Gaussian kernels to improve detection performance while maintaining computational efficiency.
Our method is primarily based on probabilistic scoring, leveraging Maximum Softmax Probability (MSP) to normalize all scores within the [0,1] interval. This probabilistic formulation enables a natural representation of OOD scores while ensuring consistent scaling across all three scenarios, thereby avoiding potential scale inconsistency issues that may arise in detection. \\
\textbf{OODD} is a test-time OOD detection method that constructs a dynamic OOD dictionary during inference, continuously collecting representative latent OOD features from test samples without requiring fine-tuning. It further combines informative inlier sampling with a priority queue-based update mechanism and a dual OOD stabilization strategy to calibrate OOD scores based on both ID-feature similarity and dynamically accumulated OOD-feature similarity, thereby improving detection performance under evolving test-time OOD scenarios.
\subsection{Training Details}
\begin{table}[ht!]
  \centering
  \small
  \begin{tabular}{lc}
    \toprule
     & ImageNet-1k  \\
    \midrule
    Fine-tune epochs & 20\\
    Batch size & 64 \\
    Initial LR & $4\times 10^{-4}$ \\
    Final LR & $4\times 10^{-5}$ \\
    LR schedule & cosine \\
    $K$ in Eq.~\ref{eq: slots logits} & 6\\
    Slot Dim. & 256 \\
    \bottomrule
    \end{tabular}%
  \caption{Configurations of \modelname.}
  \label{tab:train}%
\end{table}%
 We present the training parameters in detail, as shown in \cref{tab:train}. We employed vanilla slot attention in our implementation.

\subsection{Quantitative Results}

\begin{figure*}[!tbp]
    \centering
    \includegraphics[width=0.8\linewidth]{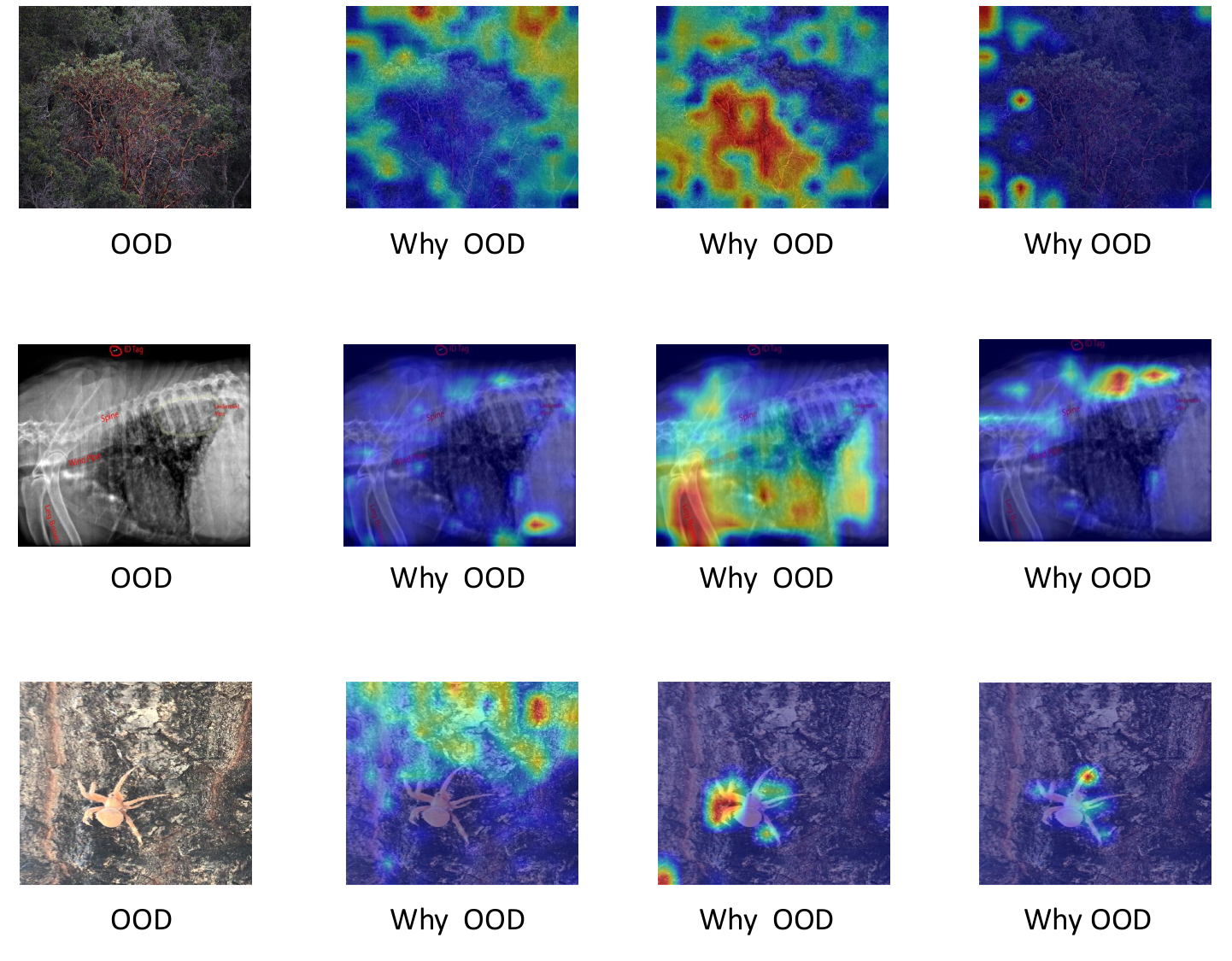}
    \caption{Attention visualization of OOD.}
    \label{fig:ood}
\end{figure*}
In this section, we visualized the regions of attention focus by plotting attention scores for each slot across different scenarios' ID images (See Fig.~\ref{fig:id}).
When no object co-occurrence (first row), the sample contains only one object besides the background. Here, each slot attends to the holistic features of the cat, leading to the prediction of \texttt{Persian cat}.
When ID object co-occurrence appears (second row), the scene is more complex. The slots attend to both the \texttt{kelpie's} features and the background, resulting in predictions of \texttt{slug} and \texttt{stick insect}. From a human visual perspective, the background indeed shares similar visual features with \texttt{slugs}. When OOD object co-occurrence appears (third row), the scene presents higher complexity with human arms intersecting with a \texttt{stingray}. Initially, the slots capture the human arm features and misidentify them as \texttt{basset}. However, the model correctly identifies the object as a \texttt{stingray} upon detecting the marine context and more comprehensive features.
The attention scores for each slot in OOD scenarios are shown in Fig.~\ref{fig:ood}. When processing OOD samples, the attention distribution across slots exhibits significantly higher dispersion.
\subsection{Visualization Results for Each Scenario}
\begin{figure}
    \centering
    \includegraphics[width=0.98\linewidth]{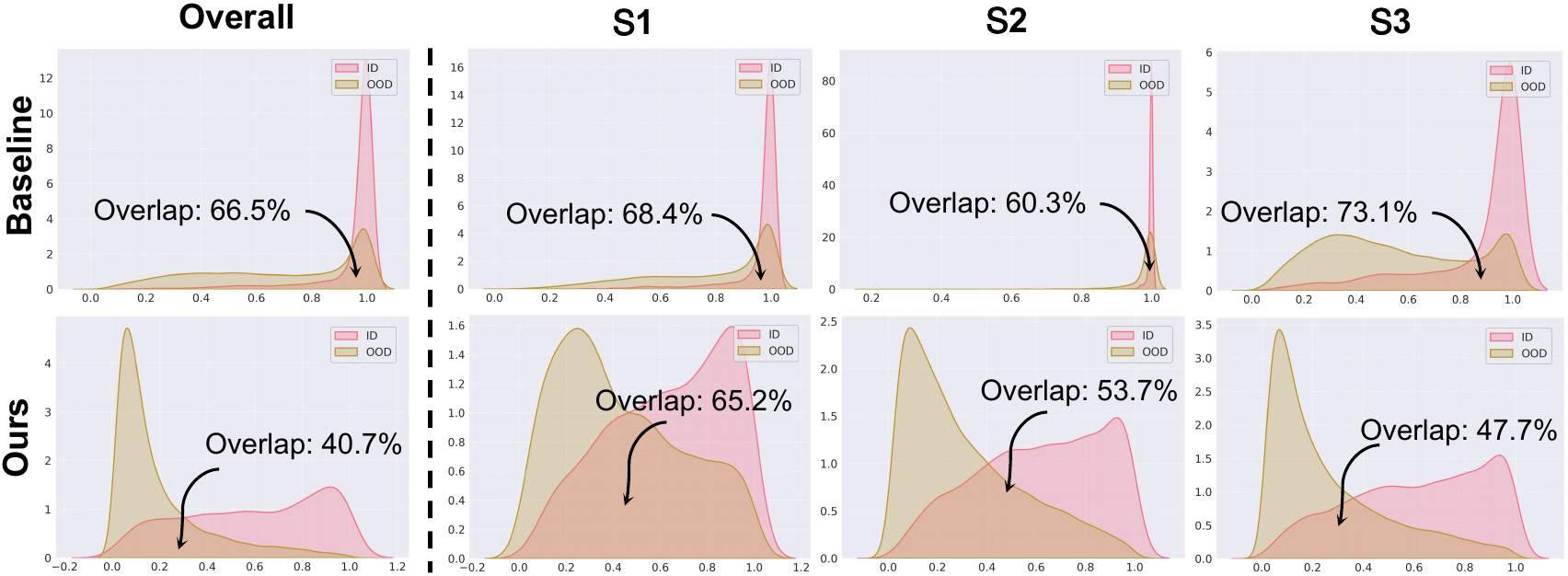}
    \caption{Score distributions for ViT model on ImageNet-200 (ID) and SSB-hard (OOD) (Left). Overall comparison of vanilla Maximum Softmax Probability (MSP) vs. OCO scores (Right). }
    \label{fig:dist}
\end{figure}

Our OCO demonstrates a clear separation between OOD distributions, as illustrated in Fig.~\ref{fig:dist}, especially in co-occurrence scenarios S2 and S3, where the advantages over traditional methods are more pronounced.
\section{Limitation}
In this section, we discuss the limitations of our method. Our model only represents object co-occurrence patterns based on slot attention. While slot attention is currently the state-of-the-art method for extracting object-centric representations, it is limited to a fixed number of slots. When the number of slots exceeds the number of objects, the representation of certain object edges may not be extracted effectively, resulting in average performance on small target objects. In the future, we can try to introduce a lightweight network to estimate the quantity, improving the dynamic number of slots, thereby making the object occurrence pattern more robust.

{
    \small
    \bibliographystyle{ieeenat_fullname}
    \bibliography{main}
}


\end{document}